\documentclass[journal,oneside]{IEEEtran}
\usepackage{mathtools}

\usepackage{graphicx}
\usepackage{lettrine}
\usepackage{booktabs}
\usepackage{amsmath}
\usepackage{enumerate}
\usepackage{bm}
\usepackage{graphicx}
\usepackage{epstopdf}
\usepackage{amsmath}
\usepackage{array}
\usepackage{amssymb}
\usepackage[square, comma, sort&compress, numbers]{natbib}
\usepackage{subfigure}
\usepackage{caption}

\usepackage{xcolor}
\usepackage{soul}

\ifCLASSINFOpdf
\else
\fi
  \usepackage[caption=false,font=footnotesize]{subfig}
\begin{document}
\title{Convolutional Sequence to Sequence \\ Non-intrusive Load Monitoring}

\author{\IEEEauthorblockN{Kunjin Chen\IEEEauthorrefmark{1},
Qin Wang\IEEEauthorrefmark{2},
Ziyu He\IEEEauthorrefmark{3}
Kunlong Chen\IEEEauthorrefmark{4},
Jun Hu\IEEEauthorrefmark{1} and
Jinliang He\IEEEauthorrefmark{1}}

\IEEEauthorblockA{\IEEEauthorrefmark{1}Department of Electrical Engineering, Tsinghua University, Beijing, China}

\IEEEauthorblockA{\IEEEauthorrefmark{2}Department of Information Technology and Electrical Engineering, ETH Z\"urich, Z\"urich, Switzerland}

\IEEEauthorblockA{\IEEEauthorrefmark{3}Department of Industrial and Systems Engineering, University of Southern California, Los Angeles, USA}

\IEEEauthorblockA{\IEEEauthorrefmark{4}Department of Electrical Engineering, Beijing Jiaotong University, Beijing, China}
}


%Telephone: (800) 555--1212, Fax: (888) 555--1212}

\maketitle
\begin{abstract}
A convolutional sequence to sequence non-intrusive load monitoring model is proposed in this paper. Gated linear unit convolutional layers are used to extract information from the sequences of aggregate electricity consumption. Residual blocks are also introduced to refine the output of the neural network. The partially overlapped output sequences of the network are averaged to produce the final output of the model. We apply the proposed model to the REDD dataset and compare it with the convolutional sequence to point model in the literature. Results show that the proposed model is able to give satisfactory disaggregation performance for appliances with varied characteristics.
\end{abstract}

\begin{IEEEkeywords}
Non-intrusive load monitoring, convolutional network, sequence to sequence learning, gated linear unit.
\end{IEEEkeywords}


% \fi
\IEEEpeerreviewmaketitle

\section{Introduction}

Non-intrusive load monitoring (NILM) refers to the technique of estimating the power demand of a single appliance from the combined demand of multiple appliances in a household measured by a single meter \cite{kelly2015neural}. It is suggested in \cite{fischer2008feedback} that electricity consumption feedback that includes appliance-specific breakdown is more likely to promote electricity conservation for residential consumers. Electricity providers can have more detailed and in-depth understanding of their customers and provide better services. Thus, both electricity consumers and electricity providers can benefit from the information provided by accurate disaggregation of whole-home power demands. 

Comprehensive reviews of various NILM methods can be found in \cite{zoha2012non, faustine2017survey}. In recent years, the success of deep neural networks (DNN) in the fields including computer vision, speech recognition, and natural language processing has gained much attention in the research community and the industry \cite{goodfellow2016deep}. As NILM is a well-defined machine learning task, researchers have been actively applying various DNN models to this task. In \cite{kelly2015neural}, the authors proposed several DNN architectures that are mainly composed of one-dimensional convolutional layers, long short-term memory (LSTM) recurrent layers, denoising autoencoders, and fully connected layers. It is believed that convolutional layers are able to extract local features of electricity consumption patterns, while LSTM layers (or other types recurrent layers) are good at modelling the temporal dependence within the sequences of electricity consumption. Some other studies focused on deep neural networks with homogenous building blocks in order to gain in-depth insights into these building blocks. For instance, in \cite{kim2017nonintrusive}, the authors propsed a network with one LSTM layer (which is in fact not a DNN). Much emphasis was placed on the fact that the long-term behavior of multi-state appliances can be properly modelled.  Another network structure that mainly consists of convolutional layers was proposed in \cite{zhang2018sequence}. The authors visualized the feature maps of the network with different input sequences in order to demonstrate the effectiveness of the model they proposed. A number of DNN models with different types of building blocks are also introduced in \cite{do2016applications}.

While the main stream practice of modelling temporal data is to leverage recurrent neural networks (RNN) \cite{graves2013speech, bahdanau2015neural}, some recent studies have shown that convolutional networks can also work well on one-dimensional data \cite{dauphin2016language, gehring2017convolutional}. We learn from these newly-developed convolutional network structures that are specially designed for one-dimensional data and adapt them to the task of NILM. As the networks consist of mainly convolutional and fully connected layers, they can be trained within a reasonably short amount of time on GPUs. We apply the proposed convolutional sequence to sequence model to a real-world dataset, and the results show that the proposed model outperforms existing models based on convolutional networks.
 
\section{Convolutional Sequence to Sequence Model for Non-intrusive Load Monitoring}
Given a whole-home power consumption sequence $\mathbf{e} = (e_1,\cdots,e_m)$, the goal of power consumption disaggregation is to obtain the power consumption sequence of the $i$th appliance, $\mathbf{h}_i = (h_{i1},\cdots,h_{im})$, where $m$ is the length of the sequences. In this paper, one convolutional sequence to sequence model is able to produce disaggregated power consumption sequences for a single appliance.

\begin{figure}[!tb]
\centering
\includegraphics[width=8.5cm]{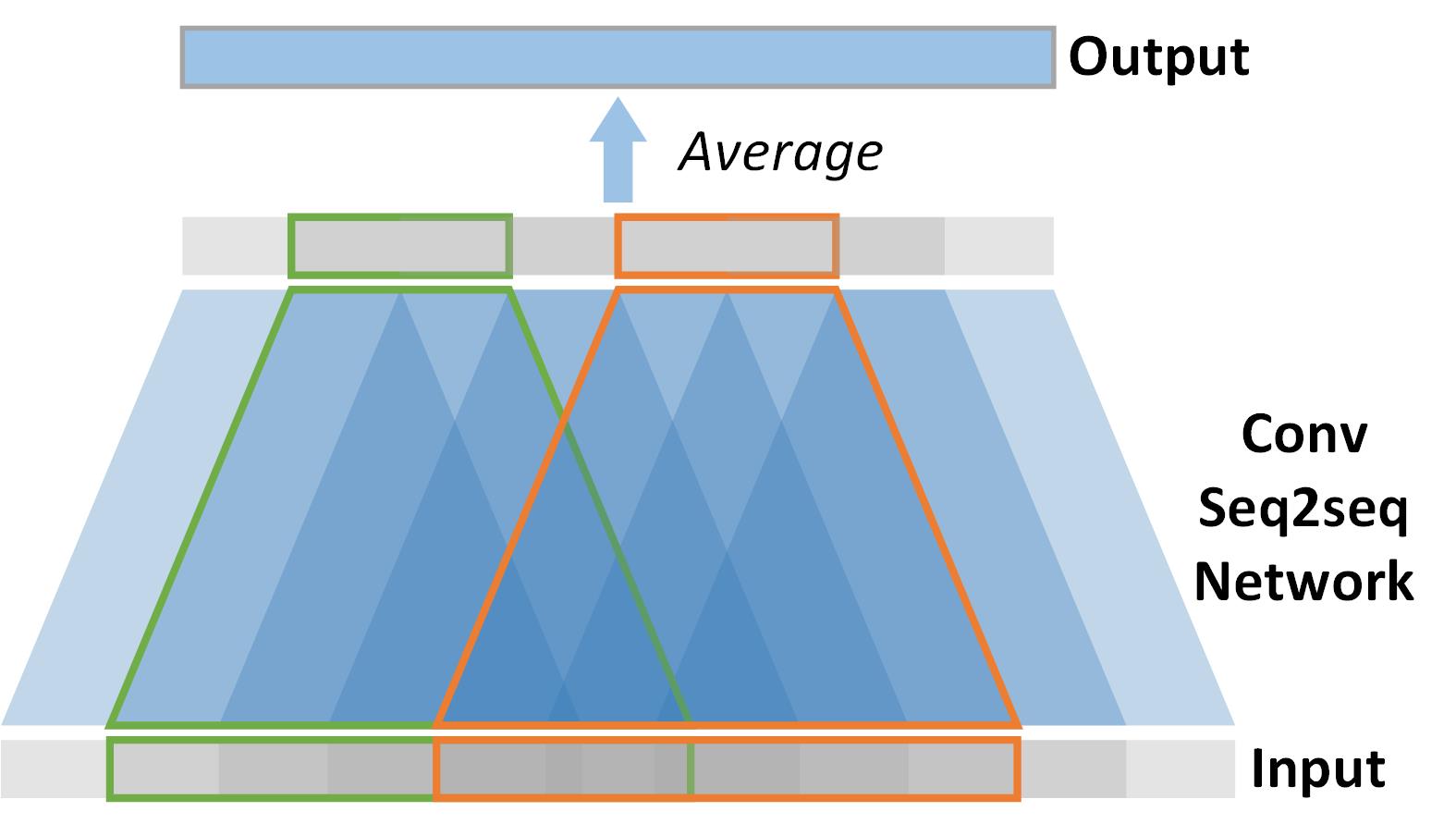}
\caption{The illustration of the convolutional sequence to sequence NILM model. The windows for input and output sequences move forward with a small step size so that multiple output windows would overlap. The final output is obtained by averaging the multiple values for each time step.}
\label{model}
\end{figure}

The illustration of the proposed model is presented in Fig. \ref{model}. Specifically, the convolutional sequence to sequence network takes the input $\mathbf{x}$ of length $l_{in}$ and maps it to the output $\mathbf{y}$ of length $l_{out}$. We set the value of $l_{in}$ several times larger than that of $l_{out}$, which helps the model perceive electricity consumptions before and after the target time range. This is different from the sequence to sequence strategies in \cite{kelly2015neural}, \cite{zhang2018sequence}, and \cite{do2016applications}, where $l_{out}$ is the same as $l_{in}$. Further, in order to make use of information from both directions of time, the window of each $\mathbf{y}$ is aligned to the center of each $\mathbf{x}$. Instead of cutting the complete output sequence into non-overlapping sections, we move the windows for the input and output sequences with a small step size of $s_{\Delta}$, which is much smaller than $l_{out}$. Thus, for the majority of the output sequence, we produce $l_{out}/s_{\Delta}$ values for each time step. We then average the values for all of the time steps and obtain the final output.

\begin{figure*}[!tb]
\centering
\includegraphics[width=14cm]{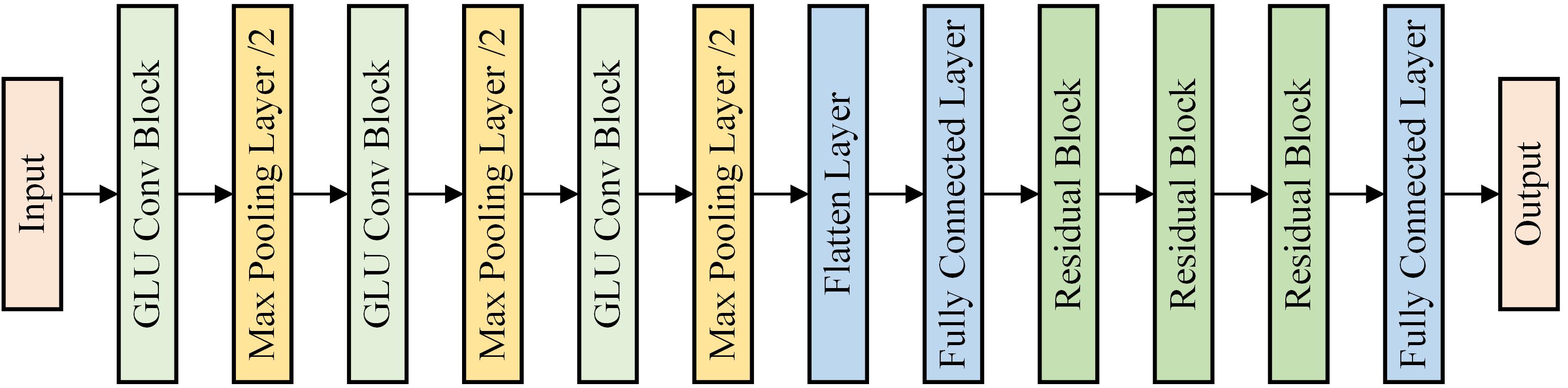}
\caption{The overall structure of the convolutional sequence to sequence network. The GLU convolutional blocks and the max pooling layers are used to process the input sequence and shrink its size. The size of the output of the fully connected layer  following the flatten layer is the same as the size of the output of the network. The residual blocks are used to refine the network's output.}
\label{network}
\end{figure*}

In Fig. \ref{network}, we demonstrate the overall structure of the convolutional sequence to sequence network. In order to process the input of the network, several gated linear unit (GLU) convolutional blocks \cite{dauphin2016language} and max pooling layers are used after the input layer. We set the pool size of the max pooling layers to 2, so that the size of the layers is divided by a factor of 2 after each pooling layer. Thus, the size of the layer after three max pooling layers becomes $l_{in}/$8. A flatten layer then reshapes the feature maps from the previous layer and produces a vector as its output, and a fully connected layer with rectified linear unit (ReLU) as its activation function is used to reduce the size of the vector to match the size of the network's output (i.e., $l_{out}$). Further, several residual blocks are added to the network to refine the output sequence. Another fully connected layer is added as the output layer of the network. For the convenience of implementation, we make sure that $l_{in}=8l_{out}$ and $l_{out}$ is an even number.

\begin{figure}[!tb]
\centering
\includegraphics[width=4cm]{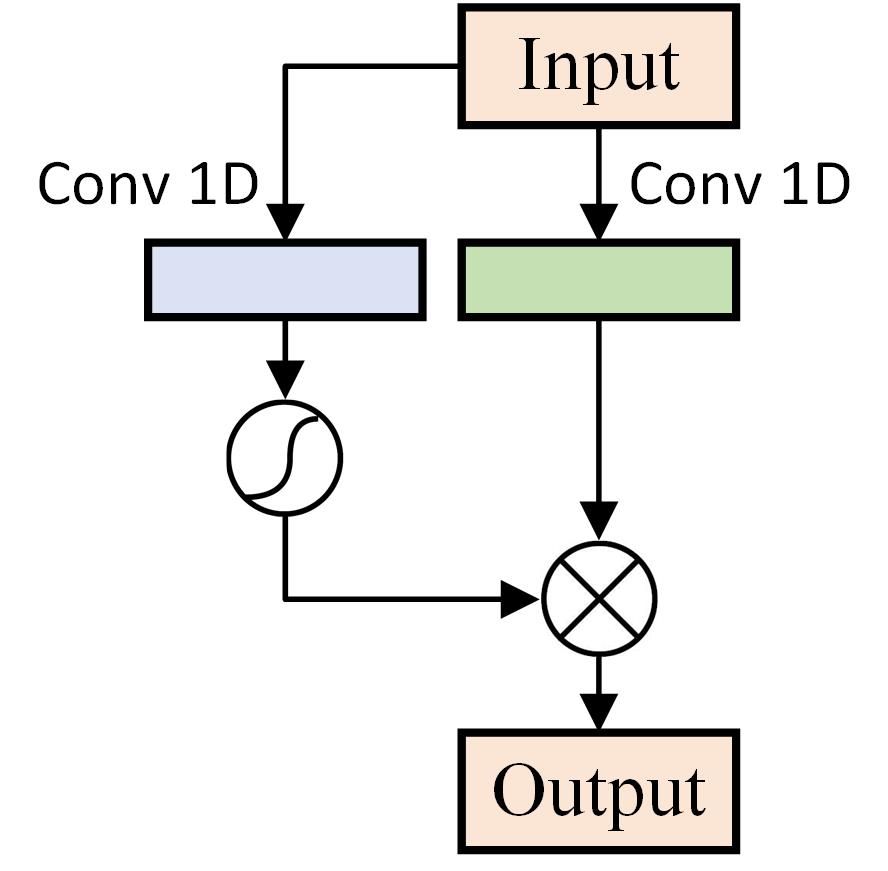}
\caption{An illustration of the GLU convolutional block. Two pathways of convolution operations are involved: the right pathway is the main pathway, and the pathway on the left is the additional pathway used to fulfill the gating mechanism.}
\label{GLU_block}
\end{figure}

The detailed structure of the GLU convolutional block, which uses GLU as its non-linearity, is shown in Fig. \ref{GLU_block}. Two streams of convolutional operations are involved, and the additional convolutional pathway (Fig. \ref{GLU_block}, left) is used to fulfill the gating mechanism of GLU. More specifically, if we denote the feature maps along the main pathway and the additional pathway as $A \in \mathbb{R}^d$, and $B \in \mathbb{R}^d$ ($d$ is the number of kernels), respectively, then the output of the convolution operation, $g$, is calculated as
\begin{equation}
g(A,B) = A \otimes \sigma(B) 
\end{equation}
where $\sigma$ is the sigmoid function, and $\otimes$ is the operation of element-wise multiplication \cite{gehring2017convolutional}. We then concatenate the outputs of each convolution operation along the input sequence and yield the output of the block.

\begin{figure}[!tb]
\centering
\includegraphics[width=3cm]{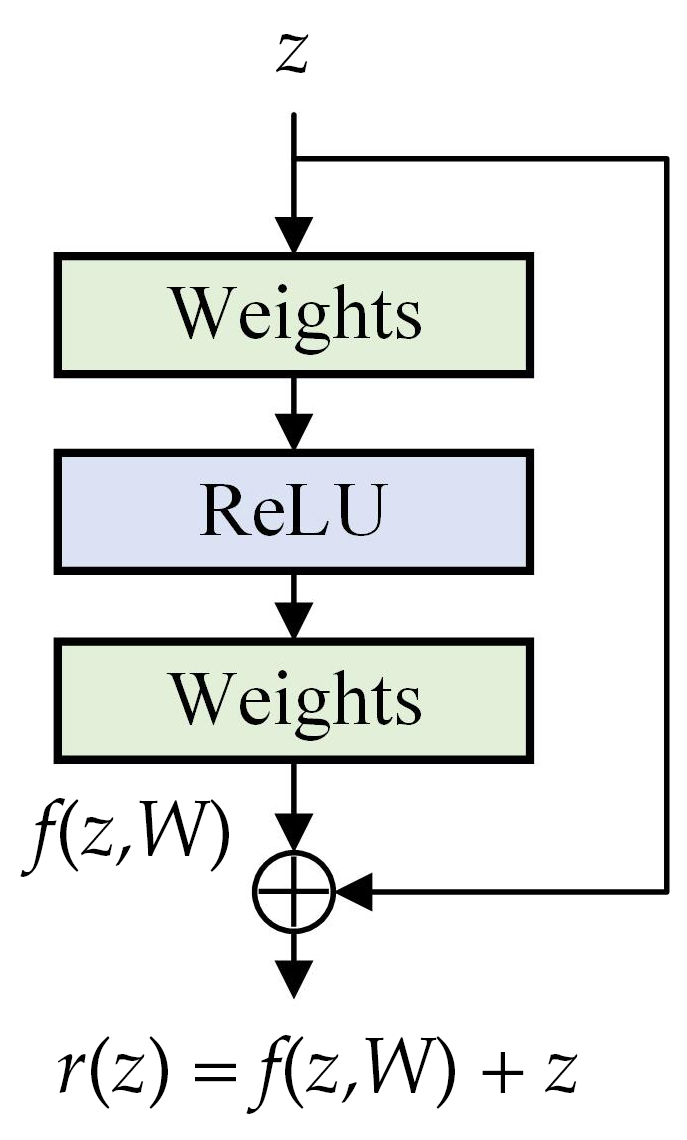}
\caption{An illustration of the residual block. A shortcut path links the input and output of the block. Two fully connected layers are used to model the residual mapping $f$.}
\label{res_block}
\end{figure}

As for the residual block (shown in Fig. \ref{res_block}), we put two fully connected layers within each block, and the ReLU non-linearity is added between the two layers. The output of the residual block is obtained by
\begin{equation}
r(\mathbf{z}) = f(\mathbf{z},W) + \mathbf{z}
\end{equation}
where $\mathbf{z}$ is the input to the block and $W$ is the set of weights and biases associated with the residual block. The adoption of residual blocks allows us to increase the learning ability of the network (i.e., increase the depth of the network) without suffering from the vanishing gradient problem \cite{he2016deep}.

\section{Experiments}

In this section, we apply the proposed model to a real-world dataset used for NILM studies and report the disaggregation results for appliances of different characteristics.

\subsection{The Dataset Used for the Experiments} 
\begin{figure}[!tb]
\centering                                                         
\subfigure[]{                   
\begin{minipage}{8cm}
\centering                                                      
\includegraphics[width=8cm]{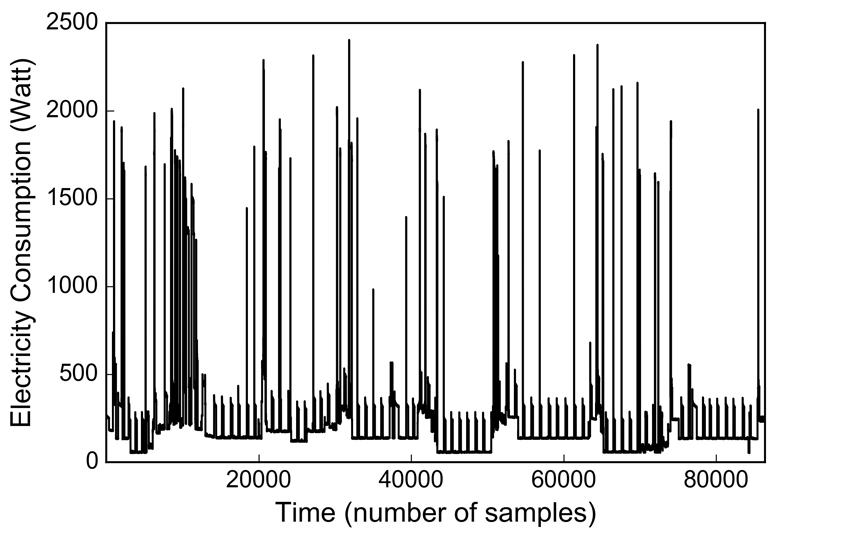}             
\end{minipage}
}
\subfigure[]{                
\begin{minipage}{8cm}
\centering                                                         
\includegraphics[width=8cm]{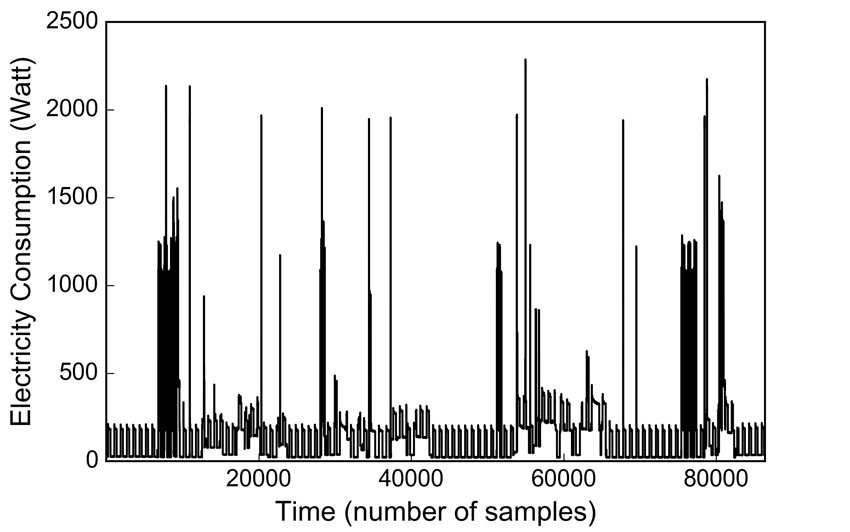}               
\end{minipage}
}
\caption{Examples of the synthesized electricity consumption time series for two houses in the REDD dataset: (a) house 1, and (b) house 2. The time range corresponds to the first few days for the data of the houses.} %                      
\label{mains}                                                      
\end{figure}

The Reference Energy Disaggregation Data Set (REDD) \cite{kolter2011redd} is used to demonstrate the effectiveness of the proposed model in this paper. The appliance-level electricity consumption values for various appliances are sampled every 3 seconds for a number of real houses. We train the models on the data of house 2 and test the models on house 1. Six types of appliances, namely, kitchen outlets, lighting devices, microwave, washer dryer, fridge, and dish washer are used to synthesize the aggregate electricity consumption time series in this paper. The data for house 2 covers a limited time range of roughly 14 days, while the time range for house 1 is more than 30 days. This is a relatively strict setting, as we require the models trained with only the data from one house to generalize to an unseen house. Examples of the synthesized household electricity consumption time series for house 1 and house 2 are plotted in Fig. \ref{mains}. While the two houses share some similar electricity consumption patterns, it is still challenging to train the model on house 2 and obtain satisfactory disaggregation results on house 1, as the appliances in the houses have different power demands and operating characteristics.

Three types of appliances, namely, fridge, lighting devices, and dish washers, are selected to be the target disaggregation appliances for the following reasons:
\begin{itemize}
\item The electricity consumption of a fridge appears to be periodic and relatively stable, thus it is easier to learn the consumption patterns for this appliance. However, as the power demands for the fridges are less than 500 Watts, the consumption of fridges can easily be masked by other appliances with high power demands. 
\item Lighting devices generally have low power consumption, thus they are hard to be separated from other appliances. In addition, the usage of lighting devices is generally more flexible and unpredictable.
\item The usage of dish washers is very sparse compared with other appliances. We can thus find out whether the proposed model is suitable for unbalanced data.  
\end{itemize}

When used as the output of the model, the power demands of fridge, lighting devices, and dish washer are divided by 500, 200, and 1400 Watts, respectively. The synthesized power demands of both houses are divided by 1000 Watts. For dish washer, we re-sample from the data of house 2 based on the operation state of the appliance so that the models are able to learn the consumption patterns effectively. More specifically, we include all the samples that correspond to the \emph{on} state of dish washer, but randomly reject samples that correspond to the \emph{off} state, so that the proportion of \emph{on} state samples is large enough for the models to be trained properly.

\subsection{The Details of the Proposed Network and the Benchmark}
The implementation details of the proposed model is as follows:
\begin{itemize}
\item The size of the input sequence is set to 800. With 3 max pooling layers with a pool size of 2, the output size of the network is 100. The step size of moving the windows along the input and output sequences is set to 5.
\item The GLU convolutional block. Both pathways have 100 kernels with a size of 4. \emph{Zero padding} is added so that the output size of the blocks is the same as the input size. 
\item The residual block. Each residual block has two fully connected layers with 50 hidden neurons. The activation function of the first layer is ReLU. 
\item The fully connected layer. Both of the two fully connected layers have 100 neurons. The first fully connected layer uses ReLU non-linearity, whereas the second fully connected layer is linear as it is the output layer of the network.
\end{itemize}

In addition, we compare the proposed model with the sequence to point convolutional network model (a seven-layer convolutional network) proposed in \cite{zhang2018sequence}. For a fair comparison, the numbers of kernels or hidden neurons for the layers within the proposed model in this paper are not tuned but set to reasonable values, such that the size of the proposed model is comparable with the model in \cite{zhang2018sequence}. Both models are trained with the Adam optimizer, and the loss is the mean absolute error (MAE) between the disaggregated and actual electricity consumptions of an appliance. The mini-batch size is set to 32. The models are implemented with \texttt{Keras} 2.0.8 \cite{chollet2015keras}, and \texttt{Tensorflow} 1.3.0 is used as the backend \cite{abadi2016tensorflow}. A Titan Xp GPU is used in order to speed up the training of the models.

\subsection{Disaggregation Results}

The MAEs of disaggregation results for different appliances are listed in Table \ref{mae}. It is seen in the table that the convolutional sequence to sequence model proposed in this paper outperforms the convolutional sequence to point model in \cite{zhang2018sequence} by a large margin. Note that the sequence to point model fails to learn the consumption patterns of lighting devices, resulting in a very large MAE with respect to the power demands of the lighting devices. For dish washer, the proposed model is able to learn the consumption patterns when the proportion of \emph{on} state samples is 10\%, while the sequence to point model requires 50\% of the samples to be of \emph{on} state (the actual proportion of \emph{on} state samples is only 1.4\%). 

\begin{table}[!t]
\renewcommand\arraystretch{1}
\centering  % 表居中
\captionsetup{justification=centering}
\caption{Mean Absolute Errors (Watts) of the Models for Fridge, Lighting Devices, and Dish Washer in House 1 of the REDD Dataset} 
\begin{tabular}{p{3.2cm} p{1.2cm} p{1.2cm} p{1.4cm}}
\toprule[1.5pt]
Model & Fridge & Lighting & Dish washer\\
\midrule[0.75pt]

Conv seq2point \cite{zhang2018sequence} & $51.5$      & $71.6$    & $32.9$ \\  
Conv seq2seq (this paper) & $\mathbf{32.0}$     &  $\mathbf{36.5}$     &  $\mathbf{12.8}$\\   

\bottomrule[1.5pt]
\end{tabular}
\label{mae}
\end{table}

\begin{figure}[!tb]
\centering                                                         
\subfigure[]{                   
\begin{minipage}{8cm}
\centering                                                      
\includegraphics[width=8cm]{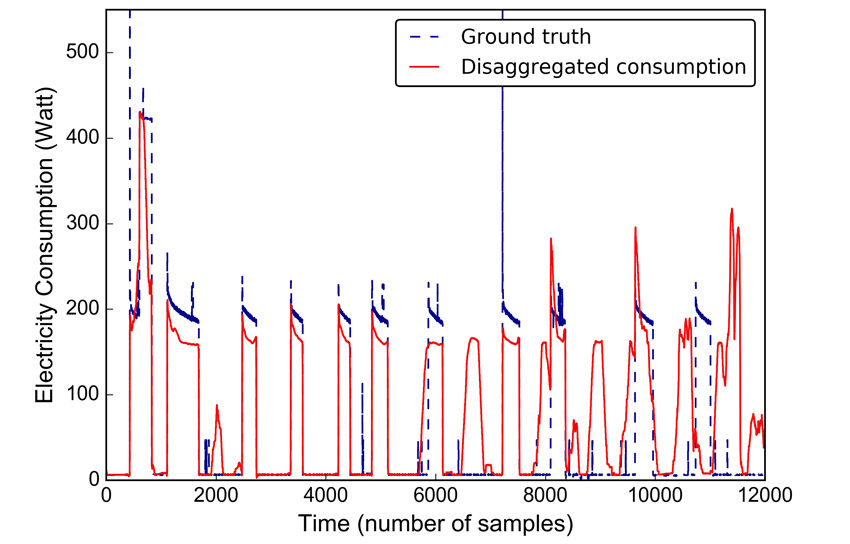}             
\end{minipage}
}
\subfigure[]{                
\begin{minipage}{8cm}
\centering                                                         
\includegraphics[width=8cm]{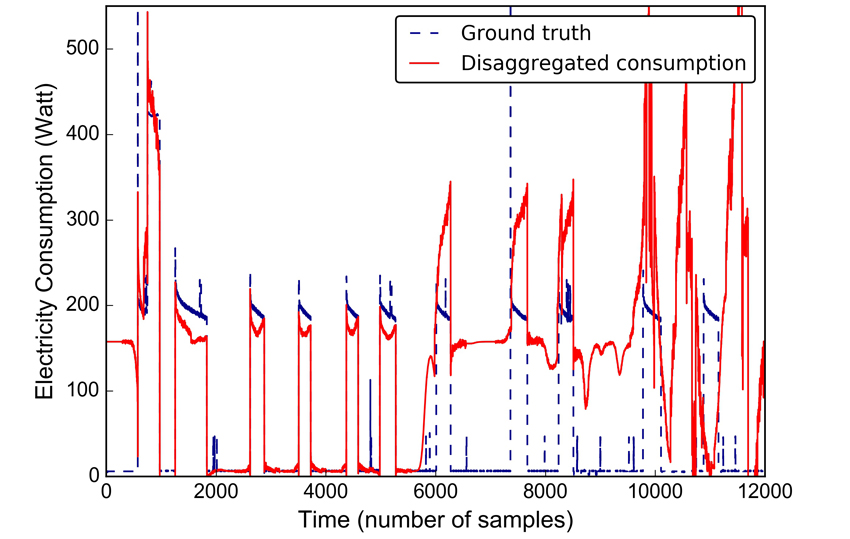}               
\end{minipage}
}
\caption{Comparison of disaggregation results for the fridge in house 1 produced by (a) the proposed model, and (b) the model in \cite{zhang2018sequence}.} %                      
\label{fridge}                                                      
\end{figure}

\begin{figure}[!tb]
\centering
\includegraphics[width=8cm]{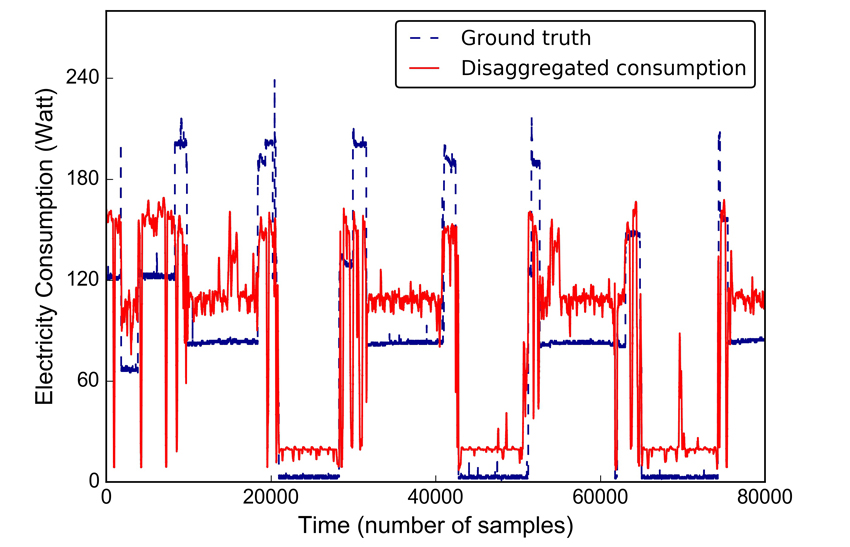}
\caption{An example of disaggregation results by the proposed model for lighting devices in house 1.}
\label{lighting}
\end{figure}

\begin{figure}[!tb]
\centering                                                         
\subfigure[]{                   
\begin{minipage}{7.8cm}
\centering                                                      
\includegraphics[width=7.8cm]{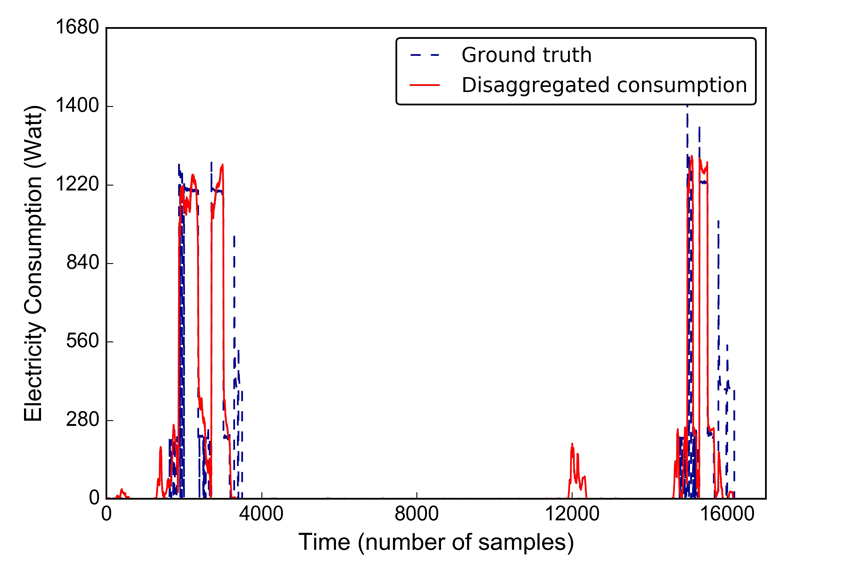}             
\end{minipage}
}
\subfigure[]{                
\begin{minipage}{7.8cm}
\centering                                                         
\includegraphics[width=7.8cm]{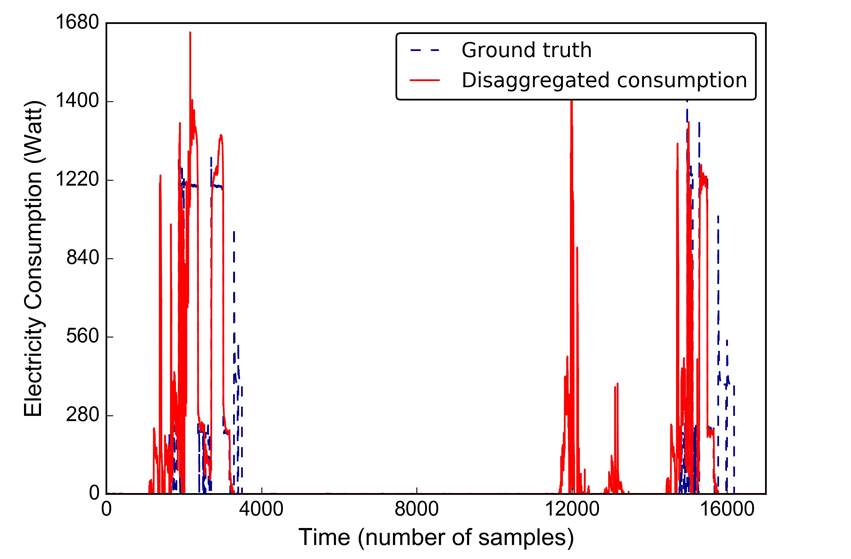}               
\end{minipage}
}
\caption{Comparison of disaggregation results for the dish washer in house 1 produced by (a) the proposed model, and (b) the model in \cite{zhang2018sequence}.} %                      
\label{dishwasher}                                                      
\end{figure}

Examples of disaggregation results for the fridge are shown in Fig. \ref{fridge}. Compared with the sequence to point model, the proposed model is more robust to highly fluctuating sections in the aggregate consumption time series (the disaggregation in the latter half of the figure is influenced by the power demands of kitchen outlets and washer dryers). 

In Fig. \ref{lighting}, we demonstrate the effectiveness of the proposed model on small power appliances (lighting devices). The power demands of different levels can be reflected by the disaggregation results. The performance of the sequence to point model is not presented, as it is unable to learn the consumption patterns and produces outputs that are all close to a small constant.

Illustrative disaggregation results for the dish washer are presented in Fig. \ref{dishwasher}. It is seen in the figure that the proposed model yields sharper shapes of demands and is less prone to produce false alarms. As a matter of fact, only one false alarm is observed for the proposed model, but the results of the sequence to point model has many spikes, partly because the model is misled by the re-sampled dataset with very high proportion of \emph{on} state samples.

\section{Conclusion and Future Work}

In this paper, we propose a sequence to sequence NILM framework based on convolutional networks. The mapping from a long sequence of aggregate power consumption to a short sequence of the power consumption of an individual appliance is facilitated by the GLU convolution blocks and max pooling layers. Further refinement of the output sequence is fulfilled by residual blocks of fully connected layers. The partial-overlapping output sequences are filtered to construct final disaggregation results. The experiments in this paper show that the proposed NILM framework outperforms existing NIML models based on convolutional neural networks. Several future paths are worth taking:
\begin{itemize}
\item The architecture of the network can be further optimized. Firstly, we may add attention mechanism to the network so that the network can learn to focus on certain parts of the aggregate electricity consumption sequence (or inputs of intermediate layers) \cite{do2016applications}. In addition, we may replace the fully connected layers within the network with convolutional networks for consistency, as the structure of a flatten layer followed by a fully connected layer partly loses the temporal relations of the points within the input sequence.
\item The way we filter the partial-overlapping output sequences can be further improved so that the model would be more robust to noisy input and complicated electricity consumption patterns. It is also of interest to integrate this process into the neural network model (e.g., the final output can be a synthesis of multiple proposals that indicate the states of the appliance \cite{ren2015faster}). 
\end{itemize}

\section*{Acknowledgement}

We are very grateful for the support of NVIDIA Corporation with the donation of the Titan Xp GPU used for this research. 

{
\small
\bibliographystyle{IEEEtran}%
\bibliography{NILM.bib}
}

%
%\begin{IEEEbiography}[Kunjin-Chen.pdf]{Kunjin Chen}
%\end{IEEEbiography}

\end{document}